\documentclass[conference,onecolumn]{IEEEtran}
\usepackage{blindtext, graphicx}
 \usepackage[table,xcdraw]{xcolor}  
\usepackage{multirow}
\usepackage{tikz,pgfplots,pgfplotstable}
\usepackage{ulem}
\usepackage{hyperref}
\usepackage{textgreek}
\pgfplotsset{compat=1.5,width=10cm}
\usepgfplotslibrary{external}
\usepackage{verbatim}
\usepackage[binary-units=true]{siunitx}
\colorlet{A1}{red!80!white}
\colorlet{B1}{orange!80!white}
\colorlet{C1}{blue!80!white}
\colorlet{A2}{red!80!black}
\colorlet{B2}{orange!80!black}
\colorlet{C2}{blue!80!black}

\usepackage{xargs}  
\usepackage[colorinlistoftodos,prependcaption,textsize=tiny]{todonotes}

\usepackage{lipsum}
\usepackage{ifthen}
\usepackage{caption}
\usepackage{pdftexcmds}
\usepackage{stringstrings}
\usepackage{multicol}
\usepackage{graphicx}
\usepackage{booktabs}

\usepackage{algorithm2e}

\usepackage{tabularx}
\usepackage{amsfonts}
\let\labelindent\relax
\usepackage{enumitem}
\usepackage{xspace}

\definecolor{bblue}{HTML}{4F81BD}
\definecolor{rred}{HTML}{C0504D}
\definecolor{ggreen}{HTML}{9BBB59}
\definecolor{ppurple}{HTML}{9F4C7C}
\newcommand{\x}{\ensuremath{\times}\xspace}

\newcommand{\xnorbin}{XNORBIN\xspace}
\newcommand{\unit}[1]{\texttt{#1}}

\newcommand{\figrefTopLevel}{Fig.~1}

\newcommand{\figrefBlockDiag}{Fig.~3}
\newcommand{\figrefKeyFigures}{Fig.~4}
\newcommand{\figrefMemory}{Fig.~5}
\newcommand{\figrefPerfAlexNet}{Fig.~6}

\newcommand{\figrefSoA}{Fig.~8}

\begin{document}
\newcommand{\andrire}[1]{\textbf{\textcolor{red}{#1}}}
\newcommandx{\info}[2][1=]{\todo[linecolor=green,backgroundcolor=green!25,bordercolor=green,#1]{#2}}
\newcommandx{\unsure}[2][1=]{\todo[linecolor=red,backgroundcolor=red!25,bordercolor=red,#1]{#2}}
\newcommandx{\change}[2][1=]{\todo[linecolor=blue,backgroundcolor=blue!25,bordercolor=blue,#1]{#2}}
\newcommandx{\improvement}[2][1=]{\todo[linecolor=Plum,backgroundcolor=Plum!25,bordercolor=Plum,#1]{#2}}
\newcommand{\hide}[1]{}
\title{\Huge\xnorbin: A 95\,TOp/s/W Hardware Accelerator for Binary Convolutional Neural Networks\\\vspace{-30pt}}

\author{\IEEEauthorblockN
{Andrawes Al Bahou\IEEEauthorrefmark{1}, Geethan Karunaratne\IEEEauthorrefmark{1}, Renzo Andri, Lukas Cavigelli, Luca Benini}
\IEEEauthorblockN{Integrated Systems Laboratory, ETH Zurich, Zurich, Switzerland} 
\IEEEauthorblockA{\IEEEauthorrefmark{1}\small These authors contributed equally to this work and are listed in alphabetical order.}
}
\maketitle\vspace{10pt}
\begin{abstract}
\normalsize
Deploying state-of-the-art CNNs requires power-hungry processors and off-chip memory. This precludes the implementation of CNNs in low-power embedded systems. Recent research shows CNNs sustain extreme quantization, binarizing their weights and intermediate feature maps, thereby saving 8-32\x memory and collapsing energy-intensive sum-of-products into \textit{XNOR}-and-\textit{popcount} operations.

We present \xnorbin, an accelerator for binary CNNs with computation tightly coupled to memory for aggressive data reuse. Implemented in UMC\,65nm technology \xnorbin achieves an energy efficiency of 95\,TOp/s/W and an area efficiency of 2.0\,TOp/s/MGE at 0.8\,V.
\end{abstract}

\IEEEpeerreviewmaketitle

\section{Introduction \& Related Work}

The recent success of convolutional neural networks (CNNs) have turned them into the go-to approach for many complex machine learning tasks.

Computing the forward pass of a state-of-the-art image classification CNN requires $\approx20$\,GOp/frame (1 multiply-accumulate corresponds to 2\,Op) and access to 20M-100M weights~\cite{canziani}. Such networks are out of reach for low-power (mW-level) embedded systems and are typically run on W-level embedded platforms or workstations with powerful GPUs. 
Hardware accelerators are essential to push CNNs to mW-range low-power platforms, where state-of-the-art energy efficiency at negligible accuracy loss is achieved using binary weight networks \cite{yoda}. For maximum energy efficiency, it is imperative to store repeatedly accessed data on-chip and limit any off-chip communication. This puts stringent constraints on the CNNs (weights and intermediate feature map size) that can fit the device, limiting applications and crippling accuracy. Binary neural networks (BNNs) use bipolar binarization (+1/-1) for both the model weights and feature maps, reducing overall memory size and bandwidth constraints by $\approx 32\times$, and simplifying the costly sum-of-products computation to mere XNOR-and-popcount operations while keeping an acceptable accuracy penalty relative to full-precision networks~\cite{rastegari}, also thanks to iterative improvement through multiple BNN convolutions \cite{NIPS2017_6638}. 

In this work, we present \xnorbin, a hardware accelerator targeting fully-binary CNNs to address both the memory and computation energy challenges. The key operations of BNNs are 2D-convolutions of multiple binary (+1/-1) input feature maps and binary (+1/-1) filter kernel sets, resulting in multiple integer-valued feature maps. This convolutions can be formulated as many parallel XNOR-and-popcount operations with the potential for intensive data reuse. The activation function and the optional batch normalization can then be collapsed to a re-binarization on a pre-computed per-output feature map threshold value and can be applied on-the-fly. An optional pooling operation can be enabled after the re-binarization. 
\xnorbin is complete in the sense that it implements all common BNN operations (e.g. those of the binary AlexNet described in \cite{rastegari}) and supports a wide range of hyperparameters. Furthermore, we provide a tool for automatic mapping of trained BNNs from the PyTorch deep learning framework to a control stream for \xnorbin.

\section{Architecture}
A top-level overview of the chip architecture is provided in \figrefTopLevel. 
The \xnorbin accelerator sequentially processes each layer of the BNN. 
\\
\textbf{Data flow:} 
To support up to $7\times 7$ kernel sizes, the processing core of \xnorbin is composed of a cluster (\figrefBlockDiag) of 7 BPUs (Basic Processing Units), where every BPU includes a set of 7 \unit{xnor\_sum} units. These units calculate the XNOR-and-popcount result on 16\,bit vectors, containing values of 16 feature maps at a specific pixel. The outputs of all 7 \unit{xnor\_sum} units in a BPU are added-up, computing one output value of a 1D convolution on an image row each cycle. On the next higher level of hierarchy, the results of the BPUs are added up to produce one output value of a 2D convolution (cf. \figrefBlockDiag). Cycle-by-cycle, a convolution window slides horizontally over the image. 
The resulting integer value is forwarded to the DMA controller, which includes a near-memory compute unit (CU). The CU accumulates the partial results by means of a read-add-write operation, since the feature maps are processed in tiles of 16. After the final accumulation of partial results, the unit also performs the thresholding/re-binarization operation (i.e. activation and batch normalization). When binary results have to be written back to memory, the DMA also handles packing them into 16\,bit words.
\\
\textbf{Data re-use/buffering:} \xnorbin comes with three levels of memory and data buffering hierarchy. \hide{An overview is given in \figrefMemory. }1) At the highest level, the main memory stores the feature maps and the partial sums of the convolutions. This memory is divided into two SRAM blocks, where one serves as the data source (i.e. contains the current layer's input feature maps) while the other serves as data sink (i.e. contains the current output feature maps/partial results), and switching their roles as layer changes. These memories are sized such that they can store the worst-case layer of a non-trivial network (implemented: 128\,kbit and 256\,kbit to accommodate binary AlexNet), thus avoid the tremendous energy cost of pushing intermediate results off-chip. Additionally, a \textit{parameter buffer} stores the weights, binarization threshold, and configuration parameters and is sized to fit the target network, but may be used as a cache for for an external flash memory storing these parameters. Integrating this accelerator with a camera sensor would allow to completely eliminate off-chip communication. 
2) On the next lower level of hierarchy, the row banks are used to buffer rows of the input feature maps to relieve pressure on the SRAM. Since these row banks need to be rotated when shifting the convolution window down, they are connected to the BPU cluster through a crossbar. 
3) The crossbar connects to the working memory inside the BPUs, the \textit{controlled shift registers} (CSRs, cf.~\figrefBlockDiag), for the input feature maps and the filter weights. These are shifted when moving the convolution window forward. All the data words in the CSRs are accessible in parallel and steered to the \unit{xnor\_sum} units. 
\\
\textbf{Scalability:} \xnorbin is not limited to $7\times 7$ convolutions. It can be configured to handle any smaller filter sizes down to $1\times 1$. However, the size of BNN's largest pair of input and output feature maps (pixels $\times$ number of maps for both) has to fit into the main memory (i.e. 250\,kbit for the actual implementation of \xnorbin). Furthermore, any convolution layer with a filter size larger than $7\times 7$ would need to be split into smaller convolutions due to the number of parallel working BPUs, \unit{xnor\_sum} units per BPU, the number of row banks, and the size of the CSRs, thereby introducing a large overhead. 
There are no limitations to the depth of the network when streaming the network parameters from an external flash memory.

\section{Implementation Results} 
We have tested our design running the binary AlexNet model from \cite{rastegari}, which comes pre-trained on the ImageNet dataset. The throughput and energy consumption per layer are shown in \figrefPerfAlexNet. 
The results are implicitly bit-true---there is no implementation loss such as going from fp32 to a fixed-point representation, since all intermediate results are integer-valued or binary. The design has been implemented in UMC 65nm technology. Evaluations for the typical-case corner at 25\,$^{\circ}$C for 1.2V and 0.8V, have yielded a throughput of 746\,GOp/s and 244\,GOp/s and an energy efficiency of 23.3\,TOp/s/W and 95\,TOp/s/W, respectively. The system consumes 1.8\,mW@0.8V from which 69\% are in the memory, 14.6\% in the DMA and crossbar and 13\% in the BPUs. The key performance and physical characteristics are presented in \figrefKeyFigures. The implementation parameters such as memory sizes have been chosen to support BNN models up to the size of binary AlexNet.
We compare energy efficiency of \xnorbin to state-of-the-art CNN accelerators in \figrefSoA. To the best of our knowledge, this is the first hardware accelerator for binary neural networks. The closest comparison point are the FPGA-based FINN results \cite{finn} with a 168\x higher energy consumption when running BNNs. The strongest competitor is \cite{yoda}, which is a binary-weight CNN accelerator strongly limited by I/O energy, requiring 25\x more energy per operation than \xnorbin.

\section{Conclusion} \label{sec:conclusion}
Thanks to the binarization of the neural networks, the memory footprint of the intermediate results as well as the filter weights could be reduced by 8-32\x, making \xnorbin capable to fit all intermediate results of a simple, but realistic BNN, such as binary AlexNet, into on-chip memory with a mere total accelerator size of 0.54\,mm${}^2$. Furthermore, the computational complexity decreases significantly as full-precision multiplier-accumulate units are replaced by XNOR and pop-count operations. Due to these benefits---smaller compute logic, keeping intermediate results on-chip, reduced model size---\xnorbin outperforms the overall energy efficiency of existing accelerators by more than 25\x.

\bibliography{xnorbin}{}
\bibliographystyle{IEEEtran}
\clearpage
\centering
\includegraphics[width=0.92\textwidth]{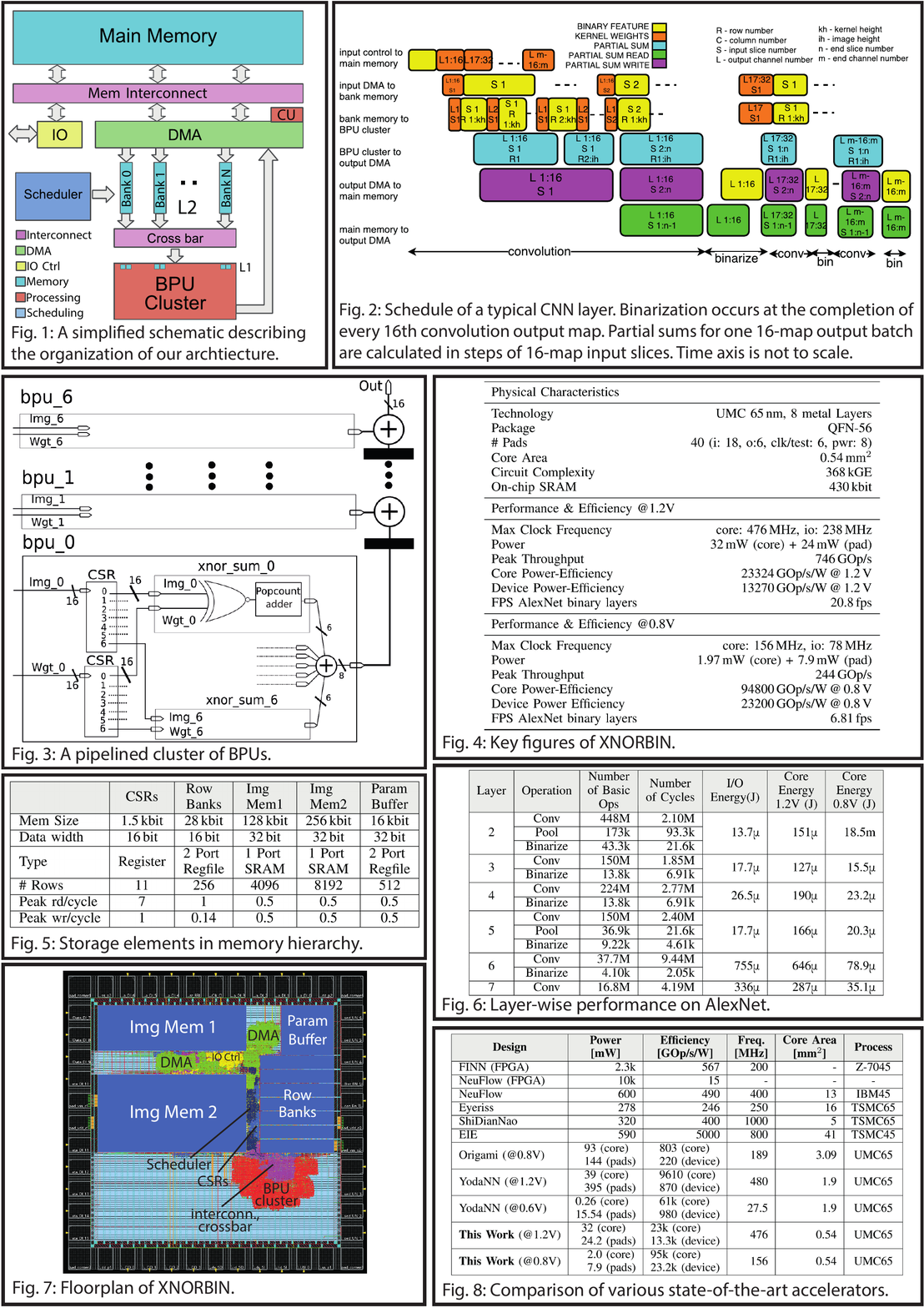}

\end{document}